\documentclass{article}

\usepackage{PRIMEarxiv}
\usepackage[T2A, T1]{fontenc}    
\usepackage[utf8]{inputenc} 
\usepackage{subfigure}
\usepackage{wrapfig}
\usepackage{hyperref}       
\usepackage{url}            
\usepackage{booktabs}       
\usepackage{amsfonts}       
\usepackage{nicefrac}       
\usepackage{microtype}      
\usepackage{lipsum}
\usepackage{listings}
\usepackage{fancyhdr}       
\usepackage{graphicx}       
\usepackage{chemformula}
\graphicspath{{images/}}     
\usepackage{authblk}
\title{eco2AI: carbon emissions tracking of machine learning models as the first step towards sustainable AI}

\author[1, 2*]{Semen Budennyy}
\author[2]{Vladimir Lazarev}
\author[1]{Nikita Zakharenko}
\author[2]{Alexey Korovin}
\author[1]{Olga Plosskaya}
\author[1, 2]{Denis Dimitrov}
\author[1]{Vladimir Arkhipkin}
\author[2, 3]{Ivan Oseledets}
\author[1]{Ivan Barsola}
\author[1]{Ilya Egorov}
\author[1]{Aleksandra Kosterina}
\author[4]{Leonid~Zhukov}

\affil[1]{Sber (AI Lab, Sber AI, ESG), Moscow}
\affil[2]{Artificial Intelligence Research Institute (AIRI), Moscow}
\affil[3]{Skolkovo Institute of Science and Technology, Moscow}
\affil[4]{Higher School of Economics University, Moscow}
\affil[*]{Corresponding author: Semen Budennyy, sanbudenny@sberbank.ru}

\begin{document}
\maketitle

\begin{abstract}

The size and complexity of deep neural networks continue to grow exponentially, significantly increasing energy consumption for training and inference by these models. We introduce an open-source package \textit{eco2AI}\footnote{Source code for \textit{eco2AI} is available at \url{https://github.com/sb-ai-lab/Eco2AI}}  to help data scientists and researchers to track energy consumption and equivalent \ch{CO2} emissions of their models in a straightforward way. In \textit{eco2AI} we put emphasis on accuracy of energy consumption tracking and correct regional \ch{CO2} emissions accounting.
We encourage  research community to search for new optimal Artificial Intelligence (AI) architectures with a lower computational cost. The motivation also comes from  the concept of AI-based green house gases sequestrating cycle with both Sustainable AI and Green AI pathways.
\end{abstract}


\section{Introduction}
While the  global ESG agenda (Environment, Social, and Corporate Governance) is guided by agreements established between countries\cite{agreement2015paris}), the development of ESG principles is happening through corporate, research, and academic standards. Many companies have started to develop their ESG strategies, allocating full-fledged functions and departments dedicated to the agenda, publishing annual reports on sustainable development, providing additional funds for research, including digital technologies and AI.

Despite growing influence of ESG agenda, it remains the problem  of transparent and objective quantitative evaluation of ESG progress in particular in environmental protection. This is of great importance for IT industry, as about one percent of the world's electricity is consumed by cloud computing, and its share continues to grow.\cite{pesce2021cloud}
Artificial Intelligence (AI) and machine learning (ML) being a big part of today's IT industry are rapidly evolving technologies with massive potential for disruption. There are number of ways in which AI and ML could mitigate environmental problems and human-induced impact. In particular, they could be used to generate and process large-scale interconnected data to learn Earth more sensitively, to predict environmental behavior in various scenarios \cite{vinuesa2020role}. This could improve our understanding of environmental processes and help us to make more informed decisions. There is also a potential for AI and ML to be used for simulating harmful activities, such as deforestation, soil erosion, flooding, increased greenhouse gases in the atmosphere, etc. Ultimately, these technologies hold great potential to improve our understanding and control of the environment.

A number of AI-based solutions are being developed to achieve carbon neutrality within the concept of Green AI. The final goal of these solutions is the reduction of Green House Gases (GHG) emissions. In fact, AI can help to reduce the effects of the climate crisis, for example, in smart grid design, developing low-emission infrastructure and modelling climate changes.\cite{dhar2020carbon} However, it is also crucial to account for generated \ch{CO2} emissions while training AI models. In fact, development of AI results into increasing computing complexity and, thereby, electrical energy consumption and resulting equivalent carbon emissions (eq. \ch{CO2}). The ecological impact of AI is a major concern that we need to account for to be aware of eventual risks. We need to ensure ML models to be environmentally sustainable, to be optimized not only in term of prediction accuracy, but also in terms of energy consumption and environmental impact. Therefore, tracking the ecological impact of AI is the first step towards Sustainable AI. Clear understanding of ecological impact from AI motivates data science community to search for optimal architectures consuming less computer resource. An explicit call to promote research on more computationally efficient algorithms was mentioned elsewhere.\cite{strubell2019energy}

\begin{figure}[!ht]
    \label{fig:scheme}
    \centering
    \includegraphics[width=0.98\linewidth]{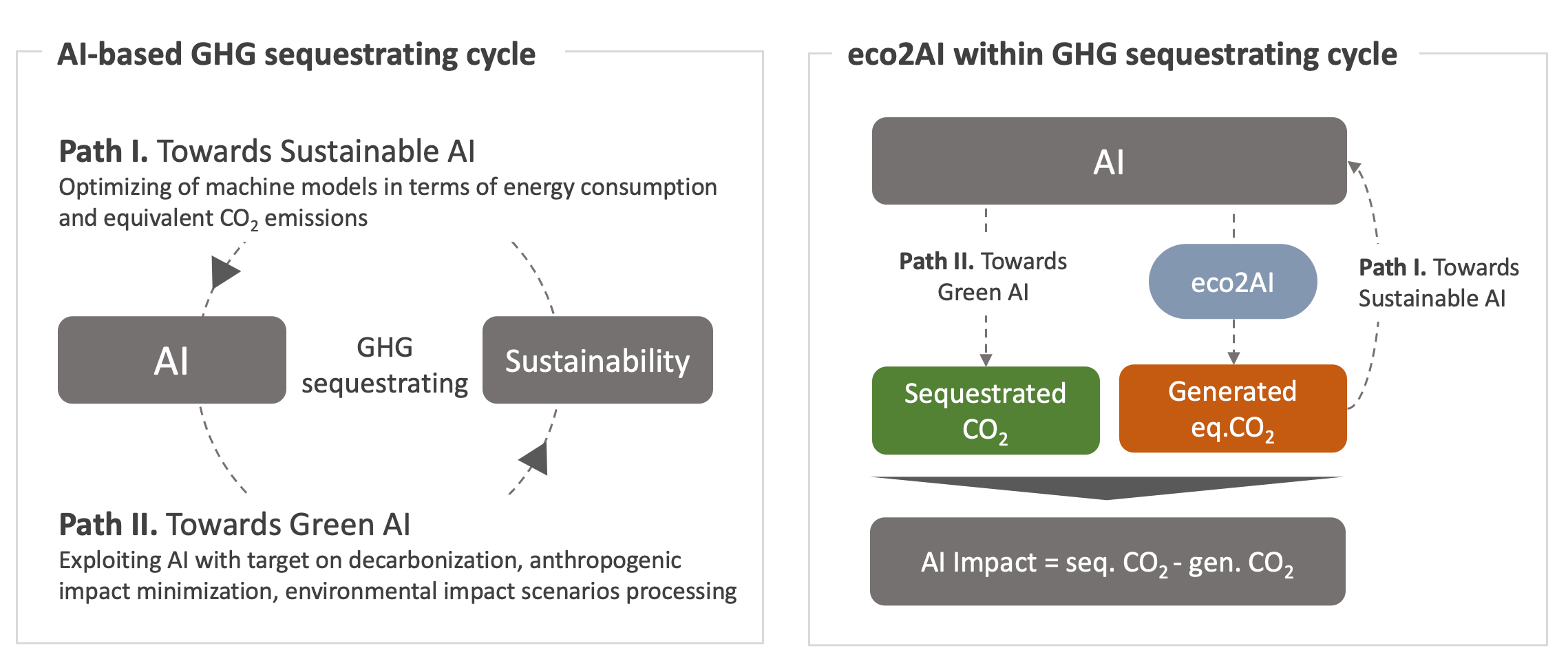}
    \caption{High-level schemes of AI-based GHG sequestrating. The left scheme corresponds to AI-based GHG sequestrating cycle. The right scheme describes the role of \textit{eco2AI} in this scheme}
\end{figure}

To summarize the previous theses, we present the concept of AI-based GHG sequestrating cycle that describes the relationship of AI with sustainability goals (Figure \ref{fig:scheme}). The request from Sustainability towards AI spawns demand for more optimized models in terms of energy consumption forming the path we named "Towards Sustainable AI". On the other hand, AI creates additional opportunities for sustainability goals' achievement, and we suggest naming this path "Towards Green AI". To understand the role of \textit{eco2AI} library in this cycle, in the right part of Figure \ref{fig:scheme} the scheme is given with paths mentioned. First, \textit{eco2AI} motivates to optimize AI technology itself. Second, if AI is aimed to sequestrate the GHG, then the total effect should be evaluated with account for generated eq. \ch{CO2}  during training sessions at least (and during model exploitation at its best). In the frame of this article, we are constrained to examining the path "Towards Sustainable AI" only (see examples in the Chapter "Experiments").

\noindent \textbf{Contribution}. The contribution of our paper is threefold: 

\begin{itemize}
\item First, we introduce \textit{eco2AI}, an open-source python library we have developed for evaluating equivalent \ch{CO2} emissions during training ML models. 
\item Second, we define the role of \textit{eco2AI} within the context of AI-based GHG sequestrating cycle concept.
\item Third, we describe practical cases where \textit{eco2AI} plays a role of efficiency optimization tracker within the context of fusion models learning.
\end{itemize}

The paper is organized as follows. In section \textbf{2} we review the existing solutions for \ch{CO2} assessment and describe their difference from our library. Section \textbf{3} presents the methodology of calculations, section \textbf{4} shows the use case of the library. Finally, in section \textbf{5} we summarize our work. The appendix section describes briefly the code usage. 

\section{Related work}
In this chapter, we describe recent practices of \ch{CO2} emissions evaluation for AI-based models. Further on, we give a brief description of the existing open-source packages, providing the summary of comparisons.

\subsection{Practice of AI equivalent carbon emissions tracking}

Since the appearance of DL models, their complexity has been increasing exponentially, doubling number of parameters every 3-4 months since 2012 \cite{AI_and_Compute_2018} and reaching more than a trillion parameters in 2022. Among the most well known models are BERT-Large (Oct 2018, 340M), GPT-2 (2019, 1.5B), T5 (Oct, 2019, 11B), GPT-3 (2020, 175B), Megatron Turing (530M), Switch Transformer (2022, 1.6T).

Data accumulation, labeling, storage, processing and exploitation  consumes a lot of  resources during their lifespan from production to disposal. The impact of such models is presented in descriptive visual map on a global scale using Amazon’s infrastructure as an example.\cite{Anatomy_of_AI} Carbon emissions are only one of footprints of such an industry but their efficient monitoring is important for passing new regulation standards and laws as well as self-regulation.\cite{henderson2020towards} 

Large-scale research was conducted focusing on quantifying the approximate environmental costs of DL widely used for NLP problems.\cite{strubell2019energy} Among the examined DL architecture, there were Transformer, ELMo, BERT, NAS, GPT-2. The total power consumption was evaluated as combined GPU, CPU and DRAM consumptions, multiplied by data center specific Power Usage Effectiveness (PUE) with default value equals 1. Sampling of CPU and GPU consumption was being queried by the vendor specialized software interface packages: Intel Running Average Power Limit and NVIDIA System Management, respectively. The conversion of energy to carbon emissions was generally carried out by multiplication of total energy consumption and carbon energy intensity. The authors estimated that carbon footprint for training BERT (base) was about 652 kg that is comparable to the footprint of the "New York <-> San Francisco" air travel per passenger.

The energy consumption and carbon footprint for the following NLP models was estimated : T5, Meena, GShard, Switch Transformer, GPT-3.\cite{patterson2021carbon} The key outcome resulted in opportunities to improve energy efficiency while training neural network models: sparsely activating DL; distillation techniques \cite{hinton2015distilling}; pruning, quantization, efficient coding \cite{han2015deep}; fine-tuning and transfer-learning \cite{dodge2020fine};  large models training in a specific region with low energy mix, exploiting cloud data centers optimized in terms of energy consumption. The authors advocated for reducing the carbon footprint by 10\textsuperscript{2}-10\textsuperscript{3} times if the mentioned suggestions had been taken into account.

\subsection{Review of open-source emission trackers}
A list of libraries have been developed to track the AI equivalent carbon footprint. Here we are focusing on describing the most widespread open-source libraries. They all have a common key goal: to monitor \ch{CO2} emissions during training models (see Table \ref{tab:1}). Having much in common with recent analogs, in \textit{eco2AI} we focused on the following: taking into account only those system processes that are related directly to models training (to avoid over-estimation); extensive and constantly updated database of regional emission coefficients (365 territorial objects are included) and information on CPU devices (3278 models).

\noindent \textbf{Cloud Carbon Footprint}\footnote{\url{https://github.com/cloud-carbon-footprint/cloud-carbon-footprint}} is an application that estimations the energy and carbon emissions of public cloud provider utilization. It measures cloud carbon and is intended to connect with various cloud service providers. It provides estimates for both energy and carbon emissions for all types of cloud usage, including embodied emissions from production, with the opportunity to drill down into emissions by cloud provider, account, service, and time period. It provides real recommendations for AWS and Google Cloud to save money and minimize carbon emissions, as well as forecasting cost savings and actual outcomes in the form of trees planted. For hyperscale data centers, it measures consumption at the service level using real server utilization rather than average server utilization. It provides a number of approaches for incorporating energy and carbon indicators into existing consumption and billing data sets, data pipelines, and monitoring systems.

\noindent \textbf{CodeCarbon}\footnote{\url{https://github.com/mlco2/codecarbon}} is a Python package for tracking the carbon emissions produced by various kinds of computer programs, from straightforward algorithms to deep neural networks. By taking into account the computing infrastructure, location, usage and running time, CodeCarbon provides an estimate of how much \ch{CO2} was produced, and gives comparisons with common modes of transportation to give an idea about scope within an order of magnitude.

\noindent \textbf{Carbontracker}\footnote{\url{https://github.com/lfwa/carbontracker}} is a tool to track and predict the energy  consumption and carbon footprint of training DL models. The package allows for a further proactive and intervention-driven approach to reducing carbon emissions by supporting predictions. Model training can be stopped at the user’s discretion if the predicted environmental cost is exceeded. Authors support a variety of different environments and platforms such as clusters, desktop computers, and Google Colab notebooks, allowing for a plug-and-play experience. \cite{anthony2020carbontracker}

\noindent \textbf{Experiment impact tracker}\footnote{\url{https://github.com/Breakend/experiment-impact-tracker}} is a  framework providing information of energy, computational and carbon impacts of ML models. It includes the following features: extraction of CPU and GPU hardware information, setting experiment start and end-times, accounting for the energy grid region where the experiment is being run (based on IP address), the average carbon intensity in the energy grid region, memory usage, the real-time CPU frequency (in Hz).\cite{henderson2020towards}

\noindent \textbf{Green Algorithms}\footnote{\url{https://github.com/GreenAlgorithms/green-algorithms-tool}} is online tool that enables a user to estimate and report the carbon footprint from computation. It integrates with computational processes and does not interfere with the existing code, while also accounting for a range of CPUs, GPUs, cloud computing, local servers and desktop computers.\cite{lannelongue2021green}

\noindent \textbf{Tracarbon}\footnote{\url{https://github.com/fvaleye/tracarbon}} is a Python library that tracks energy consumption of the device and calculates carbon emissions. It detects the location and the device model automatically and can be used as a command line interface (CLI) with predefined or calculated with the API (Application Programming Interface) user metrics.

\begin{table}
   \caption{Features of open-source trackers for equivalent \ch{CO2} emission evaluation of machine learning models} 
   
   \label{tab:1}
   \small
   \centering
   \begin{tabular}{p{0.20\linewidth} | p{0.09\linewidth}  p{0.07\linewidth}  p{0.07\linewidth} p{0.09\linewidth}  p{0.09\linewidth}  p{0.09\linewidth} p{0.09\linewidth}}
   \toprule\toprule
   \textbf{Library} & \textbf{Cloud Carbon Footprint} & \textbf{Code \linebreak Carbon} & \textbf{Carbon Tracker} & \textbf{Experimental Impact Tracker} & \textbf{Tracarbon} & \textbf{Green \linebreak Algorithms} & \textbf{eco2AI} \\ 
   \midrule
   \textbf{General information} \\
   
   Launch date & 2020 & 2020 & 2020  & 2019 & 2022 & 2021 & 2022\\
   License type  & Apache 2.0 & MIT & MIT & MIT  & Apache 2.0 & CC-BY-4.0 &  Apache 2.0 \\

    \hline
    \textbf{Carbon intensity}  & \checkmark &  \checkmark & Undefined &  \checkmark &  \checkmark &  \checkmark &  \checkmark$^{*}$ \\
   
   \hline
   \textbf{OS compatibility} \\

    Linux  & \checkmark & \checkmark & \checkmark &  \checkmark &   & \checkmark&  \checkmark\\
    Windows  & \checkmark & \checkmark & \checkmark &  &  &\checkmark &  \checkmark\\
   MacOS  & \checkmark & \checkmark & \checkmark & \checkmark  & \checkmark & \checkmark&  \checkmark\\
   
   \hline
   \textbf{Hardware compatibility} \\
   
   RAM  & \checkmark & \checkmark & \checkmark & \checkmark & \checkmark & & \checkmark \\
   CPU  & \checkmark   & \checkmark  & Undefined & \checkmark  & \checkmark  & \checkmark  & \checkmark $^{**}$  \\
   GPU  & \checkmark & \checkmark & \checkmark & \checkmark & \checkmark & \checkmark & \checkmark \\
 
   \hline
   \textbf{Supplementary} \\
   Data encryption${^*}{^*}{^*}$ &  &  &  &  &  &  & \checkmark \\
   WEB interface  & \checkmark & \checkmark &  &  & & \checkmark &  \\
   
   \bottomrule
   \end{tabular}
   \footnotesize\raggedright{$^*$ account for 365 territorial objects including regional data for Australia\cite{carbonfootprint2021, australia2021}, Canada\cite{carbonfootprint2021,Canada_2021}, Russia\cite{rosstat, fedstat} and USA\cite{carbonfootprint2021, USA2020}} \\
   \footnotesize\raggedright{${^*}{^*}$ \textit{eco2AI} database includes data on 3278 models of CPU for Intel and AMD} \\
   \footnotesize\raggedright{${^*}{^*}{^*}$ beneficial in scenarios where the authenticity of results is required } \\

\end{table}

\section{Methododology}

The methodology covers the following: calculation of electric energy consumption, extracting of emission intensity coefficient and conversion to equivalent \ch{CO2} emissions. Each part is described below. 

\subsection{Electric energy consumption}
The energy consumption of the system can be measured in Joules (J) or kilowatt-hours (kWh) - unit of energy equal to one kilowatt of power sustained for one hour. The task is to evaluate energy contribution for each  hardware unit.\cite{henderson2020towards} We focused on the GPU, CPU and RAM energy evaluation for their direct and most significant impact on the ML processes. While examining CPU and GPU energy consumption we aware of importance of tracking terminating processes but we neglect those tail effect for its relatively small impact to the total energy consumption. The storage (SSD, HDD) is also an energy consuming process but we do not take it into account as it has lack of direct relationship with running process (it is rather an issue of permanent data storage process).

\noindent \textbf{GPU}. 
The \textit{eco2AI} library is able to detect NVIDIA devices. A Python interface for GPU management and monitoring functions was implemented within the \textit{Pynvml} library. This is a wrapper for the NVIDIA Management Library which detects most of NVIDIA GPU devices and tracks the number of active devices, names, memory used, temperatures, power limits and power consumption of every detected device. Correct functionality of the library requires CUDA installation on a computing machine. The total energy consumption of all active GPU devices $E_{GPU}$  (kWh) equals to product of power consumption of GPU device and its loading time: $E_{GPU} = \int_{0}^{T} P_{GPU}(t) dt$, where $P_{GPU}$ is total power consumption of all GPU devices defined by \textit{Pynvml} (kW), $T$ is GPU devices loading time (h). If the tracker does not detect any GPU device, then GPU power consumption is set equal to zero.

\noindent \textbf{CPU}. The python modules \textit{os} and \textit{psutil} were used to monitor CPU energy consumption. To avoid overestimation, \textit{eco2AI} takes into account the current process running in the system related only to model training. Thereby, the tracker takes percentage of CPU utilization and divides it by number of CPU cores, obtaining CPU utilization percent. We realized currently the most comprehensive  database containing  3279 unique processors for Intel and AMD models. For each CPU model name provided thermal design power (TDP) which is equivalent power consumption at long-term loadings. The total energy consumption of all active CPU devices $E_{CPU}$ (kWh) is calculated as a product of the power consumption of the CPU devices and its loading time $E_{CPU} = TDP \int_{0}^{T} W_{CPU}(t) dt$, where $TDP$ is equivalent CPU model specific power consumption at long-term loading (kW), $W_{CPU}$ is the total loading of all processors (fraction). If the tracker can not match any CPU device, the CPU power consumption is set to constant value equal to 100 W\cite{maevsky2017evaluating}. 

\noindent \textbf{RAM}. Dynamic random access memory devices is important source of energy consumption in modern computing systems especially when significant amount data should be allocated or processed. However, accounting of RAM energy consumption is problematic as its power consumption is strongly depends if data is read, written or maintained. In \textit{eco2AI} RAM power consumption is considered proportional to amount of allocated power by current running process calculated as follows: $E_{RAM} = 0.375\int_{0}^{T} M_{RAM_{i}}(t) dt$, where $E_{RAM}$ - power consumption of all allocated RAM (kWh), $M_{RAM_{i}}$ is allocated memory (GB) measured via \textit{psutil} and 0.375 W/Gb is estimated specific energy consumption of DDR3, DDR4 modules\cite{maevsky2017evaluating}.


\subsection{Emission intensity}

There is variation in emissions among countries due to different factors, such as climate change, geographical position, economic development, fuel use and technological advancement. To account for regional dependence we use the emission intensity coefficient $\gamma$ that is a weight in kilogram of emitted \ch{CO2} per each megawatt-hour (MWh) of electricity generated by the particular power sector of the country. The emission intensity coefficient is totally defined by regional energy mix, or $\gamma = \sum_{i} f_i e_i$, where $i$ is an index related to the $i$-th energy source (e.g. coal, renewable, petroleum, gas, etc.), $f_i$ is a fraction of the $i$-th energy source for specific region, $e_i$ is its emission intensity coefficient. Consequently, the higher fraction of renewable energy is, the less the total emission intensity coefficient we expect. In the opposite case, high fraction of hydrocarbon energy resources implies a higher value of emission intensity coefficient. Thereby, the emission intensity varies significantly depending on the regional allocation (see Table \ref{tab:2}).

\begin{table}[!ht]
   \caption{Emission intensity coefficients for selected regions} 
   
   \label{tab:2}
   \small
   \centering
   \begin{tabular}{p{0.15\linewidth} | p{0.12\linewidth}  p{0.12\linewidth}  p{0.12\linewidth} p{0.09\linewidth}  p{0.2\linewidth}}
   \toprule\toprule
   \textbf{Country} & \textbf{ISO-Alpha-2 code} & \textbf{ISO-Alpha-3 code} & \textbf{UN M49 code} & \textbf{Emission coefficient, kg/MWh} \\ 
   \midrule
   
   \textbf{Canada} & CA & CAN & 124 & 120.49 \\
   \textbf{France} & FR & FRA & 250 & 67.53 \\
   \textbf{India} & IN & IND & 356 & 625.57 \\
   \textbf{Paraguay} & PY & PRY & 600 & 23.92 \\
   \textbf{Zambia} & ZM & ZMB & 894 & 120.78 \\
   \bottomrule
  \end{tabular}
  
\end{table}

The \textit{eco2AI} library includes permanently enriched and maintained database of emission intensity coefficients for 365 regions based on the public available data in 209 countries\cite{Global_Electricity_Review_2022} and also regional data for such countries as Australia\cite{carbonfootprint2021, australia2021}, Canada\cite{carbonfootprint2021,Canada_2021}, Russia\cite{rosstat, fedstat, minprirody500} and the USA\cite{carbonfootprint2021, USA2020}.
Currently, this is the largest database among the trackers reviewed, which allows to enrich the higher precision of energy consumption estimations.

The database contains the following data: country name, ISO-Alpha-2 code, ISO-Alpha-3 code, UN M49 code and emission coefficient value. As an example, the data for selected regions is presented in Table \ref{tab:2}. The \textit{eco2AI} library automatically defines a  user calculation facility country by IP and extracts its emission intensity coefficient. If the coefficient is not extracted for some reason, it is set to 436.5 kg/MWh, which is global
average.\cite{Global_Electricity_Review_2022}

\subsection{Equivalent carbon emissions}

Finally, the total equivalent emission value as an AI carbon footprint $CF$ (kg) generated during models learning is defined by multiplication of total power consumption from CPU, GPU and RAM by emission intensity coefficient $\gamma$ (kg/kWh) and $PUE$ coefficient: $CF = \gamma \cdot PUE \cdot (E_{CPU} + E_{GPU} + E_{RAM}).$ Here, $PUE$ is power usage effectiveness of data center required if the learning process is run on cloud. PUE is the optional parameter with default value 1. It is defined manually in the \textit{eco2AI} library.

\section{Experiments}

In the current chapter, we present experiments of tracking equivalent \ch{CO2} emissions using \textit{eco2AI} while training of Malevich (ruDALL-E XL 1.3B) \cite{Habr_ruDALLE} and Kandinsky (ruDALL-E XXL 12B)\footnote{\url{ https://github.com/sberbank-ai/ru-dalle}} models. Malevich and Kandinsky are large multimodal models\cite{gusak2022survey} with 1.3 billion and 12 billion parameters correspondingly capable of generating arbitrary images from a russian text prompt that describes the desired result.

We present results for fine-tuning Malevich and Kandinsky on the Emojis dataset\cite{shonenkov2021emojich} and for training of Malevich with optimised variation of GELU\cite{hendrycks2016gaussian} activation function. Training of the last mentioned version of Malevich allows us to consume about $10\%$ less power and, consequently, produce less equivalent \ch{CO2} emissions.

\subsection{Fine-tuning of multimodal models}

In this section we present \textit{eco2AI} use cases for monitoring fine-tuning of Malevich and Kandinsky  models characteristics (e.g., \ch{CO2}, kg; power, kWh) on the Emojis dataset. Malevich and Kandinsky are multi-modal pre-trained transformers that learn the conditional distribution of images with by some string of text. More precisely, they autoregressively model the text and image tokens as a single stream of data (see, e.g., DALL-E \cite{Ramesh21}). These models are transformer decoders~\cite{Vaswani17} with 24 and 64 layers, 16 and 60 attention heads, 2048 and 3840 hidden dimensions, respectively, and standard GELU nonlinearity. Both Malevich and Kandinsky work with 128 text tokens, which are generated from the text input using YTTM tokenizer\footnote{\url{https://github.com/VKCOM/YouTokenToMe}}, and 1024 image tokens, which are obtained encoding the input image using generative adversarial network Sber-VQGAN encoder part\footnote{\url{https://github.com/sberbank-ai/sber-vq-gan}} (it is pretrained VQGAN \cite{esser2020taming} with Gumbel Softmax Relaxation \cite{kusner2016gans}). 
The dataset of Emojis\footnote{\url{https://www.kaggle.com/datasets/shonenkov/russian-emoji}} for fine-tuning contains $2749$ unique emoji icons and $1611$ unique texts that were collected by web scrapping (the difference in quantities is due to the fact that there are sets, within which emojis differ only in color, moreover, some elements are homonyms).

\begin{table}[!ht]
   \caption{Carbon emissions and power consumption of the fine-tuning of Malevich and Kandinsky models}
   
   \label{tab:3}
   \small
   \centering
   \begin{tabular}{p{0.12\linewidth} | p{0.1\linewidth}  p{0.12\linewidth}  p{0.08\linewidth} p{0.09\linewidth}  p{0.09\linewidth} p{0.09\linewidth} p{0.09\linewidth}}
   \toprule\toprule
   \textbf{Model} & \textbf{Train time} & \textbf{Power, kWh} & \textbf{\boldmath{\ch{CO2}}, kg} & \textbf{GPU} & \textbf{CPU} & \textbf{Batch Size} \\ 
   \midrule
   \textbf{Malevich} & 4h 19m & 1.37 & \textbf{0.33} & A100 Graphics, 1 & AMD EPYC 7742 64-Core & 4 \\
   \textbf{Kandinsky} & 9h 45m & 24.50 & \textbf{5.89} & A100 Graphics, 8 & AMD EPYC 7742 64-Core & 12 \\
   \hline
   \end{tabular}
\end{table}

Malevich and Kandinsky were trained in fp16 and fp32 precision correspondingly. Adam (8-bit) \cite{dettmers20218} is used for optimization in both experiments. This realization reduces the amount of GPU memory required for gradient statistics. One cycle learning rate is chosen as a scheduler with the following parameters: start learning rate (lr) ~$4\cdot10^{-7}$, max lr~$10^{-5}$, final lr~$2\cdot10^{-8}$. Models fine-tuned for $40$ epochs with warmup $0.1$, gradient clipping $1.0$, batch size $4$ for Malevich and batch size $12$ for Kandinsky, with large image loss coefficient $1000$ and with frozen feed forward and attention layers. Malevich and Kandinsky model were trained at 1 GPU Tesla A100 (16 GB) and 8 GPU Tesla A100 (80 Gb), respectively.  It is worth mentioning that distributed model training optimizer DeepSpeed ZeRO-3 \cite{DeepSpeed_ZeRO3} was used to train Kandinsky model. The source code used for fine-tuning of Malevich is available in Kaggle\footnote{\url{https://www.kaggle.com/shonenkov/emojich-rudall-e}}.
Summary of fine-tuning parameters, energy consumption results ans eq. CO2 is given in (Table \ref{tab:3}). One can note that fine-tuning of Kandinsky consume more than 17 times more than Malevich.

\begin{figure}[!ht]
    \centering
    
    \subfigure{
    \includegraphics[width=1.0\columnwidth]{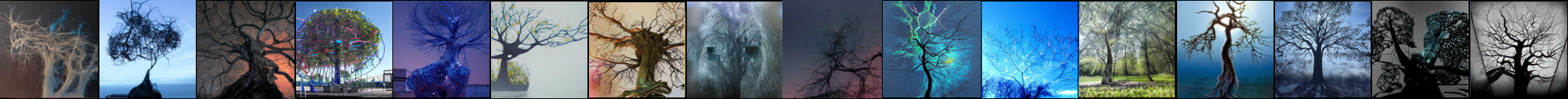}} 
    
    \subfigure{
    \includegraphics[width=1.0\columnwidth]{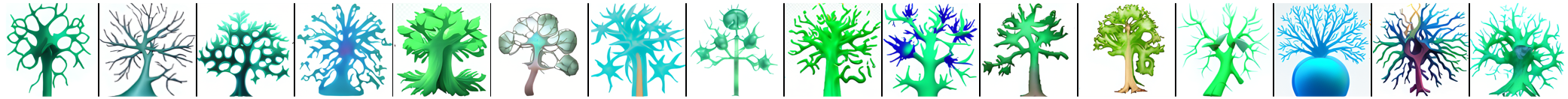}} 
    
    \caption{Images generaton of Malevich (top) vs Emojich XL (bottom) by text input ''Tree in the form of a neuron''}
    \label{fig:figure1}
\end{figure}

\begin{figure}[!ht]
    \centering
    
    \subfigure{
    \includegraphics[width=1.0\columnwidth]{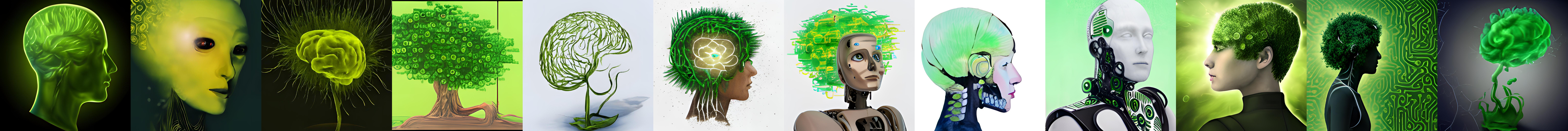}}
    
    \subfigure{
    \includegraphics[width=1.0\columnwidth]{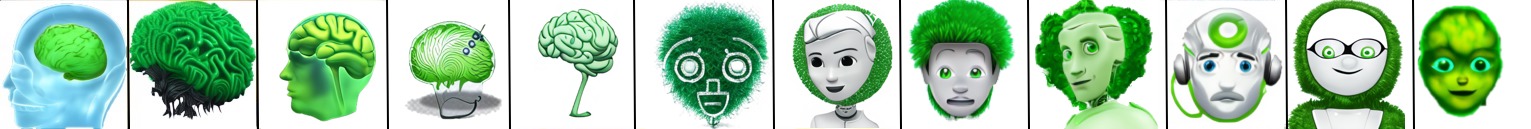}} 
     \caption{Images generation of Kandinsky (top) vs Emojich XXL (bottom) by text input ''Green Artificial Intelligence''}
    \label{fig:figure2}
\end{figure}

We have named the results of Malevich and Kandinsky fine-tuning as Emojich XL and Emojich XXL respectively. We compare the results of generation by Malevich vs by Emojich XL and by Kandinsky vs by Emojich XXL on some text inputs (see Figures \ref{fig:figure1} and \ref{fig:figure2}) to assess visually the quality of fine-tuning (how the style of generated images is adjusted to the style of emojis).

The image generation starts with a text prompt that describes the desired content. When the tokenized text is fed to Emojich, the model generates the remaining image tokens auto-regressively. Every image token is selected item-by-item from a predicted multinomial probability distribution over the image latent vectors using nucleus top-p and top-k sampling with a temperature \cite{DBLP:journals/corr/abs-1904-09751} as a decoding strategy. The image is rendered from the generated sequence of latent vectors by the decoder part of the Sber-VQGAN. 

All examples below are generated automatically with the following hyper-parameters: batch size $16$ and $6$, top-k $2048$ and $768$, top-p $0.995$ and $0.99$, temperature $1.0$, $1$ GPU Tesla A100 for Malevich (as well as Emojich XL) and Kandinsky (as well as Emojich XXL), respectively.

Thus, one can see the \textit{eco2AI} library makes it straightforward to control the energy consumption while training (and fine-tuning) large models not only on one GPU, but also on multiple GPUs, which is essential in case of using of optimisation libraries for distributed training, for example DeepSpeed ZeRO-3.

\subsection{Pre-training of multimodal models}

\begin{wrapfigure}[17]{r}{0.4\textwidth}
  \vspace{-0.2cm}
  \begin{center}
    \includegraphics[width=0.35\textwidth]{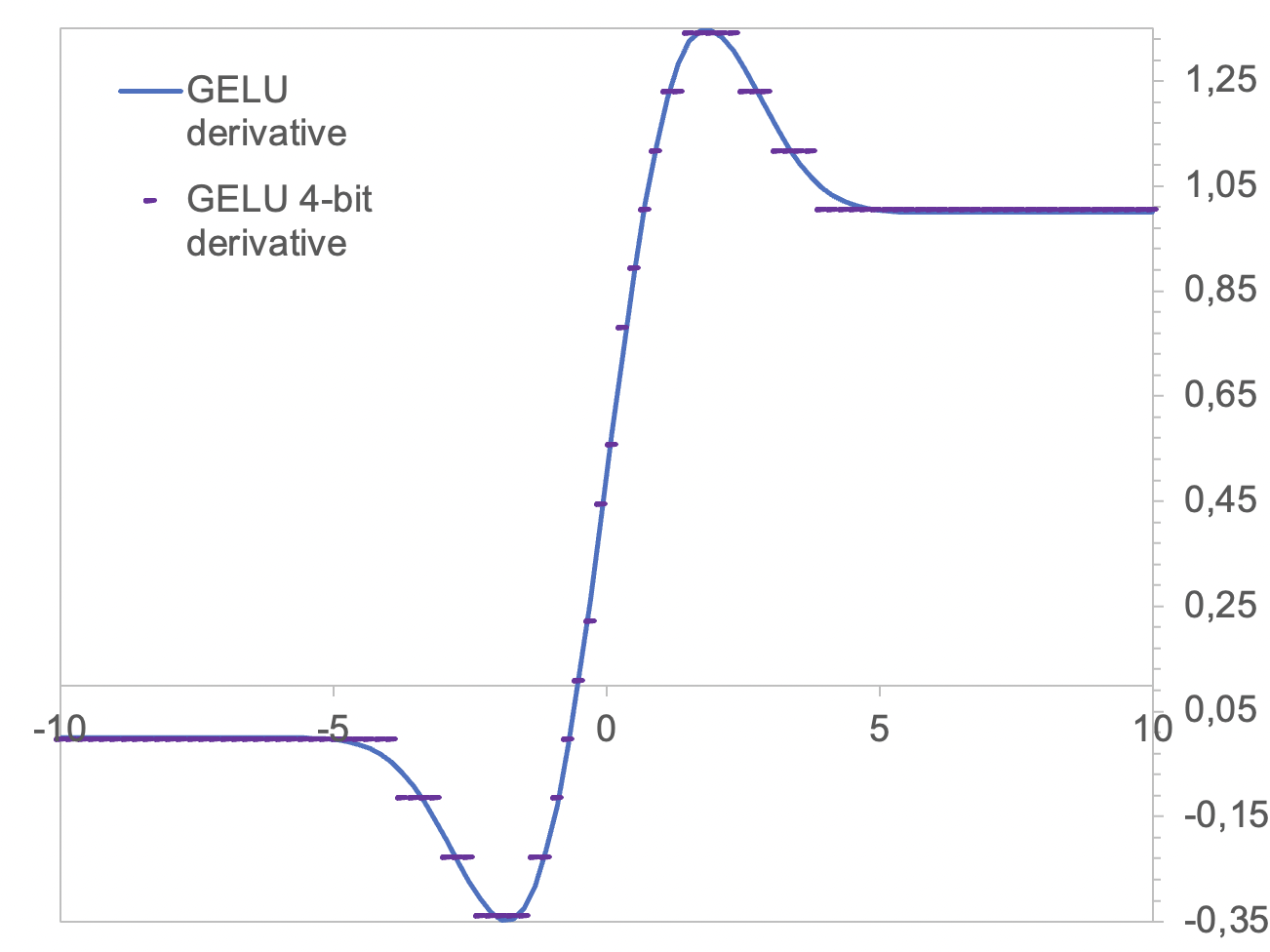}
  \end{center}
  \caption{Optimized 4-bit piecewise-constant approximation of the derivative of the GELU activation function.}
  \label{fig:figure_gelu_gelu4bit}
\end{wrapfigure}

Training large models like Malevich is highly resource demanding task. In this section we give an example of improvement its energy efficiency referring to low precision computing using 4-bit GELU activation functon as example. More precisely, we compare training of version of Malevich with regular GELU and version of Malevich with GELU 4-bit using \textit{eco2AI} library. 

GELU 4-bit \cite{novikov2022fewbit} is variation of GELU \cite{hendrycks2016gaussian} activation function that saves model gradients with 4-bit resolution thus allocating less GPU memory and spending less computational resources (see Figure \ref{fig:figure_gelu_gelu4bit}). Here we present the comparison of loss and energy efficiency Malevich model with integrated GELU and GELU 4-bit activation functions.
We used the same optimizer, scheduler and training strategy as in fine-tuning experiments. To rule out randomness, we fixed seed equls to $6 965$. Training dataset was consisted of $250 000$ samples (pairs of images and corresponding description in natural language that was balanced over the following $15$ domains: animal, nature, city, indoor, person, food, vehicle, device, tool, accessory, product, clothes, sport, art, other). Each sample was passed through the model only once with batch size $4$. Validation dataset was consisted of $5 000$ samples (pairs of images and text that have been balanced over the same domains). \textit{eco2AI} library was used to track the carbon footprint during the training in real time. 

\begin{figure}[!ht]
    \centering
    \includegraphics[scale=.52]{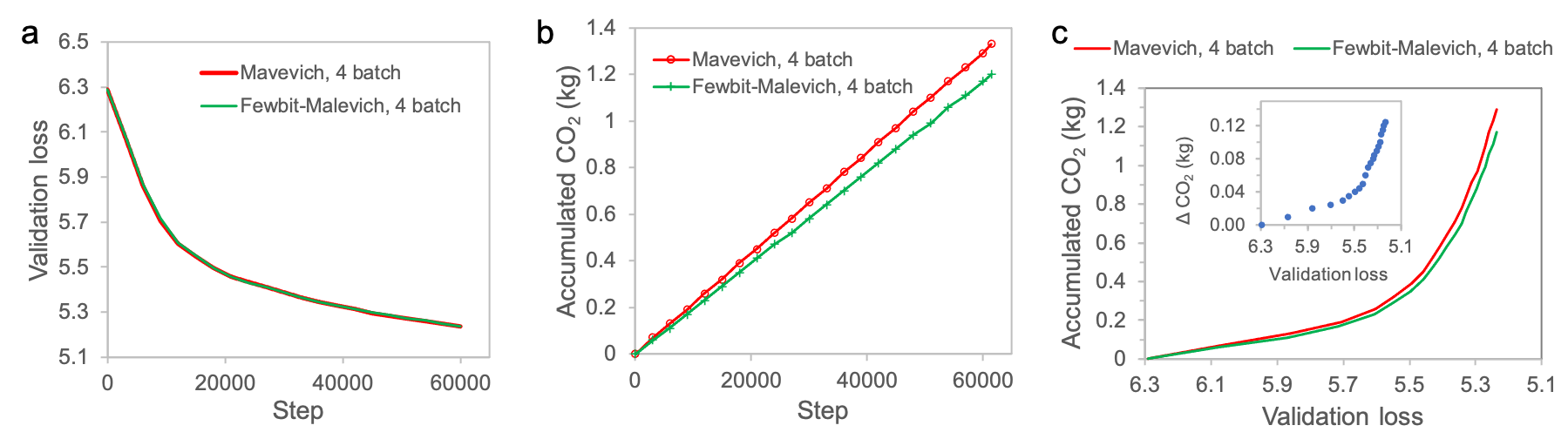}
     \caption{The comparison of GELU and GELU 4-bit activation functions integrated to Malevich model: (a)~Validation loss at every step of pre-training, (b)~Accumulated \ch{CO2} at every step of models pre-training, (c)~Accumulated \ch{CO2} for achieved validation loss of each model (the inset depicts the difference of accumulated \ch{CO2} between models)}
    \label{fig:figure4}
\end{figure}

As we can see in Figure \ref{fig:figure4}(a) validation losses of Malevich with GELU 4-bit and Malevich with regular GELU are almost the same. But GELU 4-bit is more efficient accumulating less \ch{CO2} emissions at the same training step Figure \ref{fig:figure4}(b) or achieved model loss Figure \ref{fig:figure4}(c).

As one can see in the Table \ref{tab:4} GELU 4-bit allows us to consume about $10\%$ less power and, consequently, produce less equivalent \ch{CO2} emissions.

\begin{table}[!ht]
   \caption{Carbon emissions and power consumption of the pre-trained Malevich model on $250 000$ dataset during $1$ epoch} 
   
   \label{tab:4}
   \small
   \centering
   \begin{tabular}{p{0.15\linewidth} | p{0.1\linewidth}  p{0.12\linewidth}  p{0.08\linewidth} p{0.09\linewidth}  p{0.09\linewidth} p{0.11\linewidth} p{0.09\linewidth}}
   \toprule\toprule
   \textbf{Model} & \textbf{Train time} & \textbf{Power, kWh} & \textbf{\boldmath{\ch{CO2}}, kg} & \textbf{GPU} & \textbf{CPU} & \textbf{Valid Loss } \\ 
   \midrule
   \textbf{Malevich} & 15h 23m & 5.51 & \textbf{1.33} & A100 Graphics, 1 & AMD EPYC 7742 64-Core & 5.24 \\
   \textbf{Malevich, GELU~4-bit} & 14h 5m & 4.99 & \textbf{1.20} & A100 Graphics, 1 & AMD EPYC 7742 64-Core & 5.24 \\
  \hline
  \end{tabular}
  
\end{table}

Thus, the \textit{eco2AI} library can monitor the power consumption and carbon footprint of training models in real time, helps to implement and demonstrate various memory and power optimization algorithms (such as quantization of gradients of activation functions).

\section{Conclusions}
Despite the great potential of AI to solve environmental issues, AI itself can be the source of indirect carbon footprint. In order to help AI-community to understand the environmental impact of AI models during training and inference and to systematically monitor equivalent carbon emissions in the this paper we introduced the tool \textit{eco2AI}. The \textit{eco2AI} is an open-source library capable to track equivalent carbon emissions while training or inferring python-based AI models accounting for energy consumption of CPU, GPU, RAM devices. In \textit{eco2AI} we put emphasis on accuracy of energy consumption tracking and correct regional \ch{CO2} emissions accounting due to precise measurement of process loading, extensive database of regional emission coefficients and CPU devices.

We present examples of \textit{eco2AI} usage for tracking fine-tuning of big text2image models Malevich and Kandinsky and also for optimisation of GELU activation function integrated to Malevich model. With the help of \textit{eco2AI} we demonstrated that usage of 4-bit GELU decreased equivalent \ch{CO2} emissions by about $10\%$. We expect that \textit{eco2AI} could help the ML community to pace to Green an Sustainable AI within the presented concept of AI-based GHG sequestrating cycle.

\newpage
\section*{Appendix. Usage of \textit{eco2AI} library}
The \textit{eco2AI} library is available as Python package. It is open-source, distributed under under the Apache 2.0 license\footnote{\url{https://www.apache.org/licenses/LICENSE-2.0}} and available for download and installation from PyPI  \footnote{\url{https://pypi.org/project/eco2AI/}} and one can also find its source-code on GitHub \footnote{\url{https://github.com/sb-ai-lab/eco2AI}}.

Once it is installed and imported into Python session, it will require to add start and stop code lines to frame the tracking session.
\begin{center}
\begin{lstlisting}[caption=Code integration,label=code1,language=Python]
import eco2ai

tracker = eco2ai.Tracker(project_name="YourProjectName", 
    experiment_description="training_the_<your_model>_model")

tracker.start()
<your gpu & (or) cpu calculations>
tracker.stop()
\end{lstlisting}
\end{center}

Another way to start working with the tracker is to use decorators. It allows marking any function and writing emission information in \textit{"emission.csv"} file every time when it is executed.

\begin{center}
\begin{lstlisting}[caption=using decorators,label=code2,language=Python]
from eco2ai import track

@track
def train_func(model, dataset, optimizer, epochs):
    ...
train_func(your_model, your_dataset, your_optimizer, your_epochs)
\end{lstlisting}
\end{center}

After the end of the session, all the results will be recorded in a local file \textit{"emission.csv"} or another name set by user. This file includes the following data: Project name (customized by user), Experiment description (customized by user), Start time (yyyy-mm-dd hh:mm:ss), Duration (sec), Power consumption (kWh), \ch{CO2} emission (kg), CPU name, GPU name, OS, Country.

The \textit{eco2AI} allows users to record information about training sessions in encrypted form as an extra function. This functionality is beneficial in scenarios where the authenticity of results is required. It is need to use the tracker property "\textit{encode}" to enable output encryption.

\begin{center}
\begin{lstlisting}[caption=using encrypted mode ,label=code3,language=Python]

import eco2ai

tracker = eco2ai.Tracker(
    file_name='encoded_emissions.csv',
    project_name="Test_1",
    experiment_description="testing_Eco2AI_in_encoding_mode",
    encode=True,
)

\end{lstlisting}
\end{center}

For users convenience, the \textit{eco2AI} implements the summary function. It aggregates information in the .csv file by the "project name" column. If user defines the kWh\_price argument, information about financial costs for each of the projects will be additionally calculated based on the duration time and price information provided by the user.

\begin{center}
\begin{lstlisting}[caption=using summary function ,label=code4,language=Python]
eco2ai.summary('emission.csv',kwh_price=0.117)
\end{lstlisting}
\end{center}

\newpage
\bibliographystyle{unsrt}  
\bibliography{main}

\end{document}